# Divide and Conquer: An Ensemble Approach for Hostile Post Detection in Hindi

Varad Bhatnagar, Prince Kumar, Sairam Moghili and Pushpak Bhattacharyya

Indian Institute of Technology Bombay, India
{varadhbhatnagar,princekumar,sairam,pb}@cse.iitb.ac.in

**Abstract** Recently the NLP community has started showing interest towards the challenging task of Hostile Post Detection. This paper presents our system for Shared Task @ Constraint2021 on "Hostile Post Detection in Hindi"[1]. The data for this shared task is provided in Hindi Devanagari script which was collected from Twitter and Facebook. It is a multi-label multi-class classification problem where each data instance is annotated into one or more of the five classes: fake, hate, offensive, defamation, and non-hostile. We propose a two level architecture which is made up of BERT based classifiers and statistical classifiers to solve this problem. Our team 'Albatross', scored 0.9709 Coarse grained hostility F1 score measure on Hostile Post Detection in Hindi subtask and secured 2nd rank out of 45 teams for the task [2]. Our submission is ranked 2nd and 3rd out of a total of 156 submissions with Coarse grained hostility F1 score of 0.9709 and 0.9703 respectively. Our fine grained scores are also very encouraging and can be improved with further finetuning. The code is publicly available[3].

## 1 Introduction

There was a spurt in popularity of social media in the 2010s. Thanks to high proliferation of internet and low cost, people from all age groups, education level and social status could access social media easily. Many businesses, organisations, traders, artists and freelancers are using social media to reach out to potential customers among the masses and share good/bad news instantly. This has also led to a lot of abuse and misuse[17][18]. It has been reported widely that social media has been used by political parties to shape opinions of the masses. It has been used by anti social elements to spread rumours, incite communities and stir up violence. It is seen as a cheap and highly efficient way to spread hate and disharmony. It has lead to violence, rioting, clashes and discrimination in society.[4] It has often been seen in the past, that such hostile posts have long lasting impact on the image of the individual/ group that they address[5]. Therefore, flagging hostile content on social media becomes an important task. This is a repetitive task which requires professionals to manually go through thousands of social media posts each day and flag/remove such posts. It has implications on their mental well being as well. Therefore, AI models which can flag such posts without human intervention are in demand and a hot research area right now. They will surely find immense application in the 2020s.

---

[1] https://constraint-shared-task-2021.github.io/
[2] https://competitions.codalab.org/competitions/26654
[3] https://github.com/varadhbhatnagar/Hostile-Post-Detection-in-Hindi

We aim to solve the problem of Hostile Post detection for the Hindi Language written in Devanagari Script. Hindi is the 3rd most spoken language in the world with 637 Million active speakers[6]. A sizeable chunk of these people prefer to use the Devanagari script in their interactions. Using a local language helps them connect better to other people living in the same region due to context and flavor captured by the local language. This 'connection' can be used in a positive way in times of crisis but it can also be used in a negative way to feed on some malpractices/beliefs/traditions which are common in that region. In India specific context, Hindi is understood by a large chunk of the people and it makes sense to use this language to spread hate and rumours.

The social media posts are to be classified into the following classes: non-hostile OR one or more of fake, hate, offensive, defamation. Our contributions is threefold as follows :

**1. Architecture :** We propose a two level ensemble architecture which is made up of BERT [3] based classifiers and statistical classifiers to solve this problem. We show our results on the above mentioned classes using the weighted F1 scores.

**2. Insights :** We present insights and observations that we inferred from the data. These could be used in learning better Hostile Post classification models for other languages.

**3. Error Analysis :** We analyse and try to find the reasons behind some of the errors that our model is making on the data. By ironing out these errors, there is potential to improve the accuracy even further.

## 2   Related Work

Automatic Hostile Post detection is a challenging problem in Natural Language Processing and the NLP community has recently shown great interest towards it. Kwok and Wang (2013) [10] categorized the text into binary labels i.e racist and non-racist. Their supervised approach to detect anti-black hate-speech got 76% classification accuracy on twitter data. A detailed survey on hate speech detection was done by Schmidt and Wiegand (2017) [13] where they have described the key areas for the task along with the limitation of their approach.

---

[4] https://www.washingtonpost.com/politics/2020/02/21/how-misinformation-whatsapp-led-deathly-mob-lynching-india/

[5] https://www.deccanherald.com/business/social-media-platforms-face-a-reckoning-over-hate-speech-855309.html

[6] https://en.wikipedia.org/wiki/List_of_languages_by_total_number_of_speakers#Top_languages_by_population

Ashwin Geet et al. [4] used BERT and fasttext embeddings to detect toxic speech. They performed binary and multiclass classification on twitter data. They studied two methods in which they used the word embedding in a DNN classifier and fine tuned the pre-trained BERT model. Salminen, Joni, et al. [12] performed feature analysis in their work and found out that BERT based features are most impactful for hate classification on social media platforms.

According to Waseem and Hovy [15] hate speech in the form of racist and sexist remarks are a common occurrence on social media. They have listed eleven criteria to put a tweet in the hate speech category. Sexist or racial slur, attacking a minority, seeking to silence a minority, criticizing a minority or defending xenophobia are some of their criterias. In similar work, Waseem et al. [14] studied and provided an assessment of influence of annotator knowledge on hate specch on twitter. In other work Wijesiriwardene et al. [16] showed that individual tweets are not sufficient to provide evidence for toxic behaviour instead context in interactions can give a better explanation.

An example of hostility in Hindi is to call someone *'chamcha'*, which literally means spoon in Hindi; however, the intended meaning in a hostile post could be 'sycophant'. A similar example can be found in the sentence *'Aaj konsa bakra fasaya hai?'* here *'bakra'* means Scapegoat but in Hindi it means goat.

The problem of hostility is not limited to a particular language. Notable work in this area in other languages are Haddad et al. (2020) [5] in Arabic , Jha et al. (2020) [9] in Hindi, Hossain et al. (2020) [6] in Bengali, Bohra et al. (2018) [2] in Hindi-English code mixed hate speech and Mathur et al. [11] Hindi-English code mixed offensive post .

## 3 Proposed Methodology

We have used Binary Relevance, which is a popular strategy used to solve multi-label classification problems. Here, an ensemble of single-label binary classifiers is trained, one for each class. Each classifier predicts either the membership or the non-membership of one class. The union of all classes that were predicted is taken as the multi-label output. Each binary classifier is trained separately on the dataset. So if we have 10 classes then we will have 10 different classifiers. These separate classifiers can be any model like Logistic Regression, SVM, Neural Network. There are no constraints on the type of the models to be used as a classifier.

### 3.1 Model

Our Binary Relevance model contains two levels of classifiers. At the first level, we have Non Hostile Classifier which classifies if given input is Non Hostile or not. The second level contains four other models which take only Hostile Data as input and classify whether the input is Hate, Fake, Defamation and Offensive respectively as shown in the Figure 1.

At the time of inference, the two levels will be connected, highlighted with red color line as shown in Figure 1. The predicted Hostile inputs from Non-Hostile model at the first level, is passed to the second level classifiers, where it is classified into the four hostile classes viz Hate, Fake, Defamation and Offensive. All the models except Defamation are BERT based models. The model for Defamation is SVM based. BERT models for Non-Hostile, Offensive and Fake are all same which takes just raw text and classify the input. For Hate class, there are slight modifications to a vanilla BERT model.

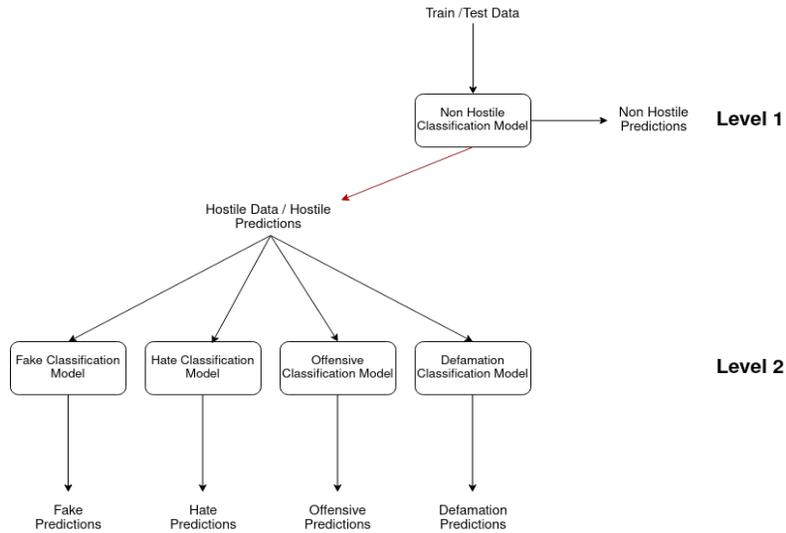

**Figure 1.** Ensemble Architecture

## 4    Implementation

### 4.1   Dataset

Definitions of the class labels (as provided to us by the organizing committee):

- **Fake News :**   A claim or information that is verified to be not true.
- **Hate Speech :**   A post targeting a specific group of people based on their ethnicity, religious beliefs, geographical belonging, race, etc., with malicious intentions of spreading hate or encouraging violence.
- **Offensive :**   A post containing profanity, impolite, rude, or vulgar language to insult a targeted individual or group.
- **Defamation :**   A mis-information regarding an individual or group.

– **Non-hostile :**   A post without any hostility.

A brief statistics of the dataset is presented in Table 1. Out of 8192 online posts, 4358 samples belong to the non-hostile category, while the rest 3834 posts convey one or more hostile dimensions.

|       | Fake | Hate | Offensive | Defamation | Total Hostile | Non-Hostile |
|-------|------|------|-----------|------------|---------------|-------------|
| Train | 1144 | 792  | 742       | 564        | 2678          | 3050        |
| Val   | 160  | 103  | 110       | 77         | 376           | 435         |
| Test  | 334  | 237  | 219       | 169        | 780           | 873         |
| Total | 1638 | 1132 | 1071      | 810        | 3834          | 4358        |

**Table 1.** Dataset statistics and label distribution

### 4.2  Experiments

**Non Hostile, Fake and Offensive Classification Model** Non Hostile model is the first level classification model in our architecture. Since there is no overlap between the Hostile and Non Hostile Classes, we are training a model to differentiate between them.

Fake and Offensive models are second level classification models in our architecture. Only the hostile samples in the dataset are used for training these models.

We are using a BERT model which has been pretrained on Hinglish[7](Mixture of Hindi and English) data. We have finetuned this model on the training data and used it as a classifier. Preprocessing the raw text data (stop word removal, etc) was leading to lower scores on the validation set because the context was being broken and language models like BERT are very sensitive to context. Hence, we have used raw text data as input to all three models. The model specifications are in Table 2.

**Hate Classification Model** This is a second level classification model in our architecture. Only the hostile samples in the dataset are used for training this model. 786 dimensional pretrained indic-bert embeddings [8], hashtags, mentions, emojis, commonly used hate words in the data and commonly used swear words in Hindi [8] are used as features in this model. Most frequently occurring mentions, emojis, words and hashtags in 'Hate' posts are one hot encoded to form the features. Threshold values (treated as hyperparameters) are used to determine the size of these vectors. We have preprocessed the raw text by removing emojis, smileys, mentions, hashtags, urls and stopwords[7]. A two layer fully connected Neural Network is trained using these features, followed by a Softmax Layer to get the output. The model specifications are in Table 3.

---
[7] https://huggingface.co/verloop/Hinglish-Bert

| Model Name | verloop/Hinglish-Bert |
|---|---|
| Architecture | 12 layer BERT followed by 1 Linear Layer |
| Features Used | Raw Text Data |
| Finetuning Epochs | 4 |
| Finetuning LR | 2e-5 |
| Finetuning Batch Size | 8 |
| Max Sentence Length | 256 |

**Table 2.** Non-Hostile, Fake and Offensive Classification Model Specifications

| Architecture | 2 layer fully connected Neural Net followed by Softmax Layer |
|---|---|
| Features Used | Emoji, Hashtag, URL, Mentions and Stopword removed data |
| Training Epochs | 10 |
| Training LR | 1e-3 |
| Training Batch Size | 4 |
| Max Sentence Length | 100 |

**Table 3.** Hate Classification Model Specifications

**Defamation Classification Model** This is a second level classification model in our architecture. Only the hostile samples in the dataset are used for training this model. We are using SVM classifier for modelling this due to low performance of BERT. BERT's lower performance can be attributed to less number of samples of this class in the training data1. FastText word embeddings, hashtags, mentions, emojis and commonly used swear words in Hindi are used as features for the classifier. Most frequently occurring mentions, emojis and hashtags in 'Defamation' posts are one hot encoded to form the features. Threshold values (treated as hyperparameters) are used to determine the size of these vectors. We have preprocessed the raw text by removing emojis, smileys, mentions, hashtags, urls and stopwords. The implementation of SVM available in python's scikit-learn library[9] has been used. The model specifications are in Table 4.

### 4.3 Binary Relevance Considerations

In a Binary Relevance Classification setting, it can so happen that a data sample is not assigned any class since all models are working in parallel. In our implementation, this happens in 71 out of 1653 test samples and we assign hate,

---

[8] https://github.com/AI4Bharat/indic-bert
[9] https://scikit-learn.org/stable/

An Ensemble Approach for Hostile Post Detection in Hindi 7

| Model Name | SVM |
|---|---|
| Parameters | Default Sklearn Parameters with class_weight = 'balanced' |
| Data Used | Emoji, Hashtag, URL, Mentions and Stopword removed data |

**Table 4.** Defamation Classification Model Specifications

offensive labels to these as it is empirically observed that the model performance increases for these two classes on doing so. This approach has been used as we were optimizing our model for the leaderboard. It can be reasoned that the results are improving because there is scope for improvement in our Hate and Offensive classifiers. Another approach could be to assign class labels based on the probability values predicted by each classifier for a particular data sample. The model would tend to overfit lesser in the latter approach as compared to the former.

## 5  Result and Analysis

Table 5 shows our results compared to the baseline model [1]. The baseline results on the validation set are presented here. Validation column shows our model results on the validation set. Validation set results are according to the evaluation script provided by the competition organisers. Ground labels of Test set was not provided during competition. Test set results mentioned in last column of Table 5 were given by organisers at the end of competition.

We can see that there is a major improvement in Coarse Grained, Fake and Offensive F1 Scores. Whereas, improvement in Defamation and Hate is much less.

|  | Baseline | Validation | Test Set |
|---|---|---|---|
| **Coarse Grained** | 0.8411 | 0.9765 | 0.9709 |
| **Defamation** | 0.4357 | 0.4951 | 0.4280 |
| **Fake** | 0.6815 | 0.8178 | 0.8140 |
| **Hate** | 0.4749 | 0.5614 | 0.4969 |
| **Offensive** | 0.4198 | 0.6108 | 0.5648 |
| **Weighted Fine Grained** | Not Given | 0.6525 | 0.6110 |

**Table 5.** F1 Score Comparison

| Class | Binary | Precision | Recall | F1 score | Support | Accuracy |
|---|---|---|---|---|---|---|
| Non-Hostile | 0 | 0.97 | 0.98 | 0.97 | 376 | 0.98 |
|  | 1 | 0.98 | 0.97 | 0.98 | 435 |  |
| Defamation | 0 | 0.91 | 0.73 | 0.81 | 305 | 0.73 |
|  | 1 | 0.39 | 0.69 | 0.50 | 74 |  |
| Fake | 0 | 0.88 | 0.87 | 0.88 | 225 | 0.85 |
|  | 1 | 0.82 | 0.83 | 0.82 | 154 |  |
| Hate | 0 | 0.78 | 0.81 | 0.80 | 270 | 0.70 |
|  | 1 | 0.48 | 0.43 | 0.46 | 109 |  |
| Offensive | 0 | 0.84 | 0.86 | 0.85 | 276 | 0.78 |
|  | 1 | 0.60 | 0.55 | 0.58 | 103 |  |

**Table 6.** Classification Report of each class. (In Binary 1-True, 0-False)

In validation data consisting of 811 inputs, 432 are classified as Non-Hostile. The remaining 379 inputs are classified independently for each class at the second level. Details of Classification report of each class is given in Table 6.

## 6   Insights

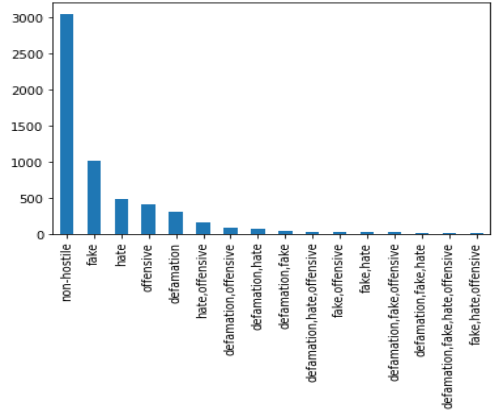

**Figure 2.** Training Set Label Distribution

There were many ideas that we tried out before arriving at the final model. Some of the insights and experiments are presented in this section.

Other strategies for solving multi label classification problems like Label Powerset were tried but the results were not very encouraging as there wasn't enough training data for every combinations of labels. As shown in Fig:2.

Since Binary Relevance ignores the correlation between labels, we tried training joint models for labels which were frequently occurring together like hate and offensive, but that did not give great results. We got f1 scores of 0.32 and 0.28 on the validation set for respective classes which is lower than the baselines. Hence, we did not pursue this approach.

Since deep learning models are very hard to explain and reason with, we initially built a statistical baseline for every class using SVM or XGBoost [10] using default hyperparameters. It helped us decide what features are important. In Defamation and Hate classifier, this knowledge was used extensively and ultimately the Defamation statistical classifier ended up outperforming the deep learning classifier on the validation set.

We inferred from the data that hateful posts generally contain language which was abusive in nature. Hence we used a list of hateful words in the Hindi language given here [9]. We added a few words to this list using the data and our own judgement.

Defamatory content usually targeted an individual, an organisation or a group of individuals based on caste, religion etc. Since data collection occurred in a specific period, we looked at the data and figured out the top entities that were being referred to in defamatory posts and used this as a feature in the classifier. We tried using emoji2vec embeddings [11] in the SVM model for Defamation class but it did not give any improvements to the F1 scores on the validation set.

## 7 Error analysis

- **Indirect Reference to Entities**

बहुत ही सुन्दर था वो
(ऐसा कहा जाता था)
पर पता नहीं कहा गायब हो गया
भाई विकास कहा हो आप

**Ground Tag**: defamation
**Predicted Tag**: offensive

It implicitly criticises the government for lack of development and progress. Understanding such indirect reference is hard for a model.

मोर को दाना खिलाने वाले से उनकी इतनी जल रही है सोचो जिस दिन बाघ को दूध पिलाने वाला आएगा तब तो इनका धुंवा निकलेगा धुंवा 🤣

---

[10] https://xgboost.readthedocs.io/en/latest/
[11] https://github.com/uclnlp/emoji2vec

- **Hard to differentiate between defamation, hate and offensive**

किसको-किसको लगता है कि राहुल गाँधी का नारा "चौकीदार चोर है" बिल्कुल सही था ! 🙊💪
😎

**Ground Tag**: hate
**Predicted Tag**: defamation,offensive

This tweet seems to be defamatory because of negativity and presence of an entity. It also seems offensive because of the word चोर.

- **How does fake work?**

किम जोंग-उन के अंतिम संस्कार का एक्सक्लूसिव वीडियो है

**Ground Tag**: fake
**Predicted Tag**: fake

How does the model predict if a tweet is fake or not? Best guess: it detect declarative sentences and tags them as fake.

कल स्वतंत्रता दिवस के अवसर पर तिरंगे को छोड़कर,नीला झंडा फहराया,पुलिस पहुंच गई और अपना डंडा लहराने लगी

**Ground Tag**: fake
**Predicted Tag**: offensive, hate

- **Annotator Bias**

चर्चित उन्नाव रेप कांड के दोषी भाजपा विधायक कुलदीप सिंह सेंगर को हाई कोर्ट से मिली जमानत। अब सवाल यह उठता है कि जिस जज ने जमानत दिया है, अगर पीड़िता उसी जज की बेटी होती, तो भी वह जज जमानत दे देता क्या? इस हैवान ने पीड़िता के पूरे खानदान को मार डाला, फिर भी उस जज को तरस नहीं आया ।

**Ground Tag**: hate, offensive
**Predicted Tag**: hate, offensive, fake and defamation

This tweets seems to be non hostile in nature, but since its talking about a heinous crime, the annotators have given it hate and offensive tags.

बिल्कुल सर। नहीं चलेगी मनमानी बीजेपी की। अपने किए वादो का 20% भी नहीं कर रही बीजेपी।

**Ground Tag** : hate.
**Predicted Tag**: defamation

## 8  Outcome

We have successfully thought of and implemented an approach to solve the problem of Hostility Detection in Hindi language. Several statistical and deep learning models have been implemented to solve the sub-problems as defined in the above sections. Our submission to the challenge is ranked 2nd and 3rd out of a total of 156 submissions with coarse grained hostility F1 score of 0.9709 and 0.9703 respectively[19].

We have presented insights and error analysis in this paper which can be used to train better models for Hostile Post detection in Hindi. Some of the work that we have done can be used as a baseline in solving similar problems for different languages.

## 9  Future Work

There are some areas in which there is scope for improvement. Using classifier chains instead of binary relevance and better feature engineering are some things that can be tried. Using out of competition data can further improve our models and there are applications in real world problems like fake news detection and flagging hostile posts on forums to which they can be applied.

---

[11] http://lcs2.iiitd.edu.in/CONSTRAINT-2021/